%% file: main.tex
\documentclass[letterpaper, 10 pt, conference]{ieeeconf}  
\IEEEoverridecommandlockouts                              

\input{irom.sty}

\usepackage{multicol}
\usepackage{times}
\usepackage{epsfig}
\usepackage{epstopdf}
\usepackage{graphicx}
\graphicspath{{Figures/}}
\usepackage{url}
\usepackage{stfloats}
\usepackage{amsmath}
\usepackage{amsfonts}   
\usepackage{amsopn}     
\usepackage{psfrag}
\usepackage{mdwtab}     
\usepackage{enumerate}
\usepackage{color}
\usepackage{accents}
\usepackage{mathtools}
\usepackage{algorithm}
\usepackage{algpseudocode}
\usepackage{adjustbox}
\usepackage{cases}
\usepackage[british]{babel}
\usepackage{hhline}
\usepackage{multirow}

\usepackage{scalerel}
\usepackage{diagbox}

\title{\LARGE \bf
FlowDrone: Wind Estimation and Gust Rejection on UAVs \\ Using Fast-Response Hot-Wire Flow Sensors
}

\author{Nathaniel Simon, Allen Z. Ren, Alexander Piqué, David Snyder, Daphne Barretto, \\ Marcus Hultmark, and Anirudha Majumdar
\thanks{N Simon, A Z Ren, A Piqué, D Snyder, M Hultmark, and A Majumdar are with the Department of Mechanical and Aerospace Engineering, and D Barretto is with the Department of Computer Science, Princeton University, Princeton, NJ, 08544.
        Emails: {\tt\small \{nsimon, allen.ren, apique, dasnyder, daphnegb, hultmark, ani.majumdar\}@princeton.edu}}%
\thanks{This work was partially funded by the Air Force Office of Scientific Research [FA9550-22-1-0020], the National Science Foundation GRFP [DGE-2039656], and Princeton University's Project X Innovation Fund. Any opinions, findings, and conclusions or recommendations expressed in this material are those of the author(s) and do not necessarily reflect the views of the AFOSR, NSF, or Princeton. M. Hultmark is co-founder and CEO of Tendo Technologies, Inc., who provided the MEMS hot-wire dies.}
}

\begin{document}

\maketitle
\thispagestyle{empty}
\pagestyle{empty}

\begin{abstract}
Unmanned aerial vehicles (UAVs) are finding use in applications that place increasing emphasis on robustness to external disturbances including extreme wind. However, traditional multirotor UAV platforms do not \emph{directly} sense wind; conventional flow sensors are too slow, insensitive, or bulky for widespread integration on UAVs. Instead, drones typically observe the effects of wind indirectly through accumulated errors in position or trajectory tracking. 
In this work, we integrate a novel flow sensor based on micro-electro-mechanical systems (MEMS) hot-wire technology developed in our prior work \cite{mst} onto a multirotor UAV for wind estimation. These sensors are omnidirectional, lightweight, fast, and accurate.
In order to achieve superior tracking performance in windy conditions, we train a `wind-aware' residual-based controller via reinforcement learning using simulated wind gusts and their aerodynamic effects on the drone. In extensive hardware experiments, we demonstrate the wind-aware controller outperforming two strong `wind-unaware' baseline controllers in challenging windy conditions. See: \href{https://youtu.be/KWqkH9Z-338}{\texttt{youtu.be/KWqkH9Z-338}}.

\end{abstract}


\input{sections/intro}
\input{sections/system}
\input{sections/sensor}
\input{sections/control}
\input{sections/results}

\input{sections/conclusion}

\bibliographystyle{IEEEtran} 
\bibliography{refs} 


\end{document}

%% file: sections/intro.tex
\section{Introduction}
Autonomous multirotor drones have the potential to transformatively impact a variety of domains including infrastructure inspection and repair, search-and-rescue operations, and aerial package delivery. Towards this end, a major challenge facing current systems is their limited ability to deal with severe wind conditions in outdoor environments, which in turn may impair reliable operations for each of the applications listed above. For example, the maximum safe wind speed for a typical multirotor drone (e.g., the DJI Phantom~\cite{dji}) is around 20 miles per hour, which roughly corresponds to a windy day at the beach. Further, the challenge posed by wind is exacerbated by the presence of complex airflow phenomena (e.g., ground and surface effects) when the drone operates in proximity to obstacles or in urban canopies.

\begin{figure}[h]
\centering
\includegraphics[width=0.48\textwidth]{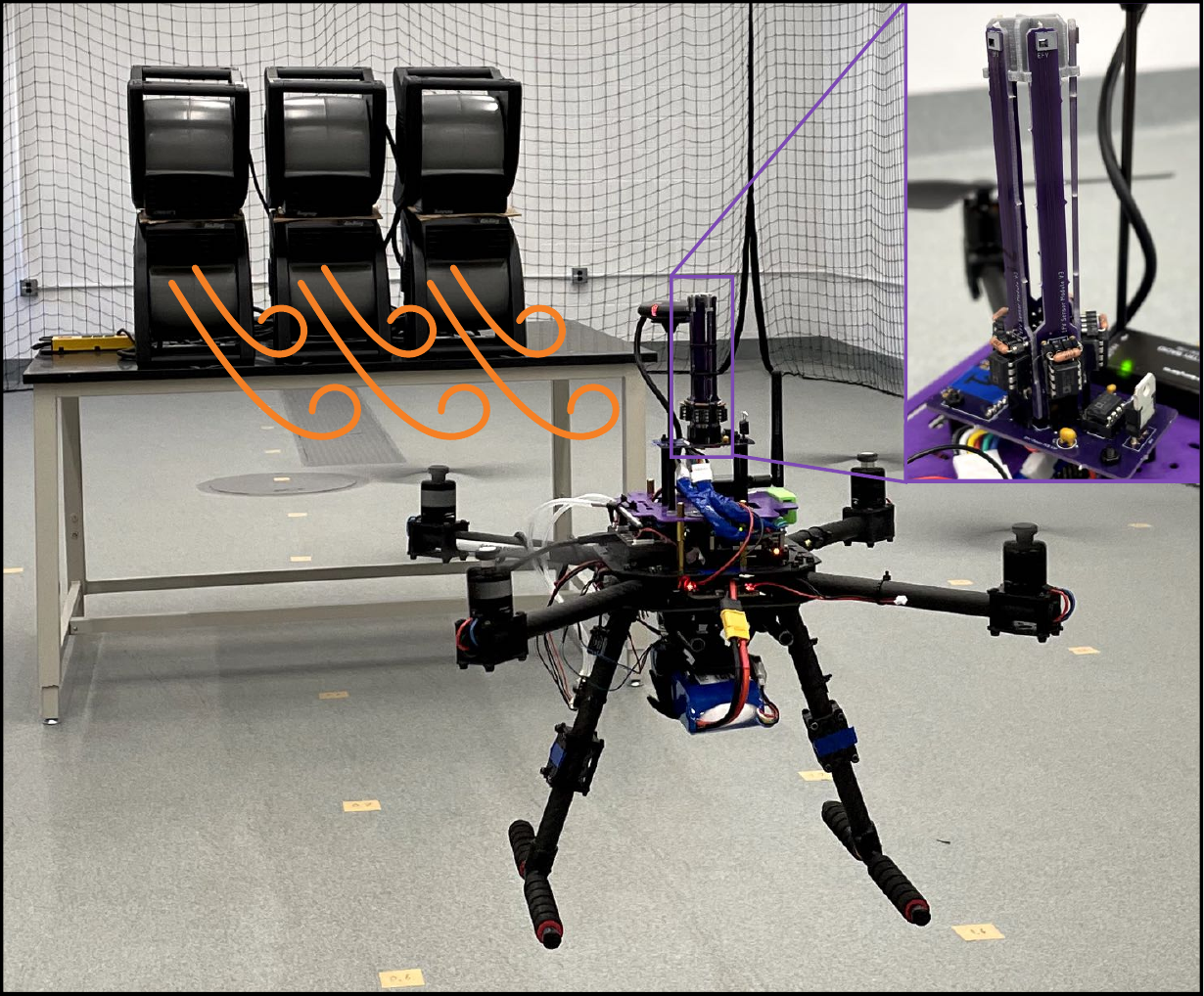}
\caption{FlowDrone incorporates the MAST (top right), an omnidirectional flow sensor leveraging MEMS hot-wires to provide fast and accurate estimates of wind magnitude and direction, enabling superior flight performance in windy conditions.}
\label{fig:anchor}
\vspace{-5mm}
\end{figure}

Currently, multirotor systems rely almost exclusively on \emph{indirect} estimates of wind for real-time control, e.g., on deviations from nominal motion measured by an inertial measurement unit (IMU), GPS, or downward-facing optical flow camera. These estimates rely on gusts impacting the drone's motion before they can be reliably estimated, and are hence inherently limited in terms of their accuracy and update rates. Existing sensors for direct wind measurement (e.g., conventional pitot tubes and hot-wires) are typically either too slow, insensitive, or lack the requisite form-factor (e.g., size and weight), and are thus rarely deployed on multirotor systems (see Sec.~\ref{sec:related-work} for an overview of existing wind sensing technology). 

The primary hypothesis behind our work is that advancements in airflow sensing technology will enable significant improvements in multirotor drone performance in severe wind conditions. In this paper, we propose the use of novel micro-electro-mechanical systems (MEMS) technology for turbulent airflow measurements \cite{fan2015nanoscale, fu2016elastic}. These sensors have three characteristics that make them ideally suited for use in multirotor control. First, they are small (on the order of a few millimeters) and lightweight (on the order of a few grams). Second, they have very low latency ($<2$ms) \cite{mst} and can operate at rates commensurate with typical drone control rates ($>500$Hz). Third, they allow accurate real-time estimation of wind magnitude and direction \cite{mst}. As such, we believe that MEMS-based airflow sensors can fill an important gap in sensing technology for multirotor drones. 


{\bf Statement of contributions.} The primary contribution of this work is to
leverage MEMS technology for turbulent airflow measurements in order to improve multirotor drone performance in high winds. To this end, we make the following three specific contributions:
\begin{itemize}[leftmargin=*]
\item {\bf FlowDrone.} We instantiate the MEMS-based flow sensing technology in the FlowDrone (Fig.~\ref{fig:anchor}): a quadrotor equipped with MEMS hot-wire airflow sensors, which enable real-time ($>500$Hz) estimates of wind magnitude and direction.
\item {\bf Wind-aware control.} We propose a reinforcement learning (RL) pipeline for training a \emph{wind-aware controller} that utilizes the wind estimate from the flow sensor. The key components of the RL training pipeline are: (i) a simulator that captures relevant aerodynamic effects, (ii) the simulation of wind gusts recorded from real experiments with additional domain randomization, and (iii) a neural network policy that takes as input a history of wind estimates along with the drone's state estimate and outputs a residual control input added to the open-source PX4 controller \cite{meier2011pixhawk}.

\item {\bf Hardware experiments.} We perform hardware experiments demonstrating the improved performance of the FlowDrone's wind-aware controller in the presence of a wind gust as compared to two strong wind-unaware baselines: (i) the widely-used PX4 controller (referred to as `baseline'), and (ii) a controller trained using RL with identical simulated winds but without the ability to access real-time wind estimates (referred to as `wind-unaware'). Over the course of 30 flights (10 per controller), each exposed to a 5~m/s gust, the mean maximum error in the direction of wind was: 0.44 m for wind-aware, 0.58 for wind-unaware, and 0.78 for baseline. This constitutes a 44\% improvement of wind-aware over baseline.
\end{itemize}

\subsection{Related Work}\label{sec:related-work}

{\bf Multirotor drone control in wind.} Current multirotor drones typically treat wind as an external disturbance and rely on a feedback controller to perform gust rejection \cite{tang2018autonomous,hoffmann2007quadrotor}, e.g., using techniques from robust and adaptive control \cite{gillula2010design,mallikarjunan2012l1,hanover2021performance}. 
Recently, techniques from adaptive control have also been combined with deep neural network representations of aerodynamic effects in order to perform multirotor control in the presence of wind \cite{o2022neural}. 

Alternative approaches to wind estimation include utilizing an extended Kalman filter with measurements from the drone's IMU (e.g., as implemented by the widely used PX4 controller \cite{usingtheeclekf}). Such wind estimates can then be utilized by a feedback controller for gust rejection \cite{schiano2014towards}. 

The approaches mentioned above rely on \emph{indirect} estimates of wind (e.g., by observing the drone's motion as measured by its IMU or optical flow sensor). In this work, we seek to leverage sensors that directly measure wind with adequate accuracy and frequency to improve multirotor drone control. 

{\bf Conventional airflow sensors.} One of the most commonly used sensors for measuring airspeed on fixed-wing platforms is the pitot-static tube \cite{anderson2007fund}.
However, they suffer from high lag and typically have settling times on the order of seconds \cite{bailey2013obtaining}, making them unreliable for unsteady flow measurements. Moreover, conventional one-dimensional pitot-static tubes do not provide estimates of flow direction, and are in fact insensitive to changes in direction of up to $15^\circ$-$20^\circ$ \cite{beck2010aerodynamics}. This renders them unsuitable as an omnidirectional flow sensor. Another conventional airflow sensor is the hot-wire anemometer \cite{bruun1995hot}. While these sensors have a significantly faster response (on the order of milliseconds or even microseconds \cite{bailey2010turbulence}), their form factor is not well-suited for multirotor applications. Specifically, they typically require large, heavy, and expensive circuitry, and are also fragile due to the unshielded microscale wire \cite{huang2021investigation}. 

{\bf Airflow sensors for multirotor drones.} Motivated by the challenges with conventional airflow sensors, there have been efforts to develop custom airflow sensing technology for multirotor drones. For example, flow speed (but not direction) can be estimated by measuring the deflection of a single sensing element \cite{sundin2021soft, zahran2018new}. The whisker-like  sensors used in \cite{tagliabue2020touch} can resolve airflow direction from deflection using multiple sensing elements; however, \cite{tagliabue2020touch} does not provide a characterization of the temporal characteristics (lag and frequency) of the sensors. Differential pressure probes have also been used for resolving the wind vector \cite{yeo2015onboard, bruschi2016wind}. However, these have relatively high errors in angle estimation (12$^\circ$ error in \cite{bruschi2016wind}), and are based on principles used by conventional airflow sensors (which suffer from high latency and low temporal resolution). Another sensing modality uses fast-response multi-hole pressure probes (MHPPs) \cite{prudden2017anemometer}, which yield wind magnitude and direction estimates of 1~m/s and 5$^\circ$ respectively. However, these sensors are unable to measure outside a 90$^\circ$ cone of acceptance, and also utilize a large probe whose length is that of the drone itself \cite{prudden2018measuring}. In this work, we utilize recently developed MEMS-based hot-wire sensors, which afford a small form factor (millimeter scale), fast response ($>500$Hz), and accurate wind speed and direction estimation. 

%% file: sections/system.tex
\section{System Overview}

\begin{figure}
\centering
\includegraphics[width=0.54\textwidth]{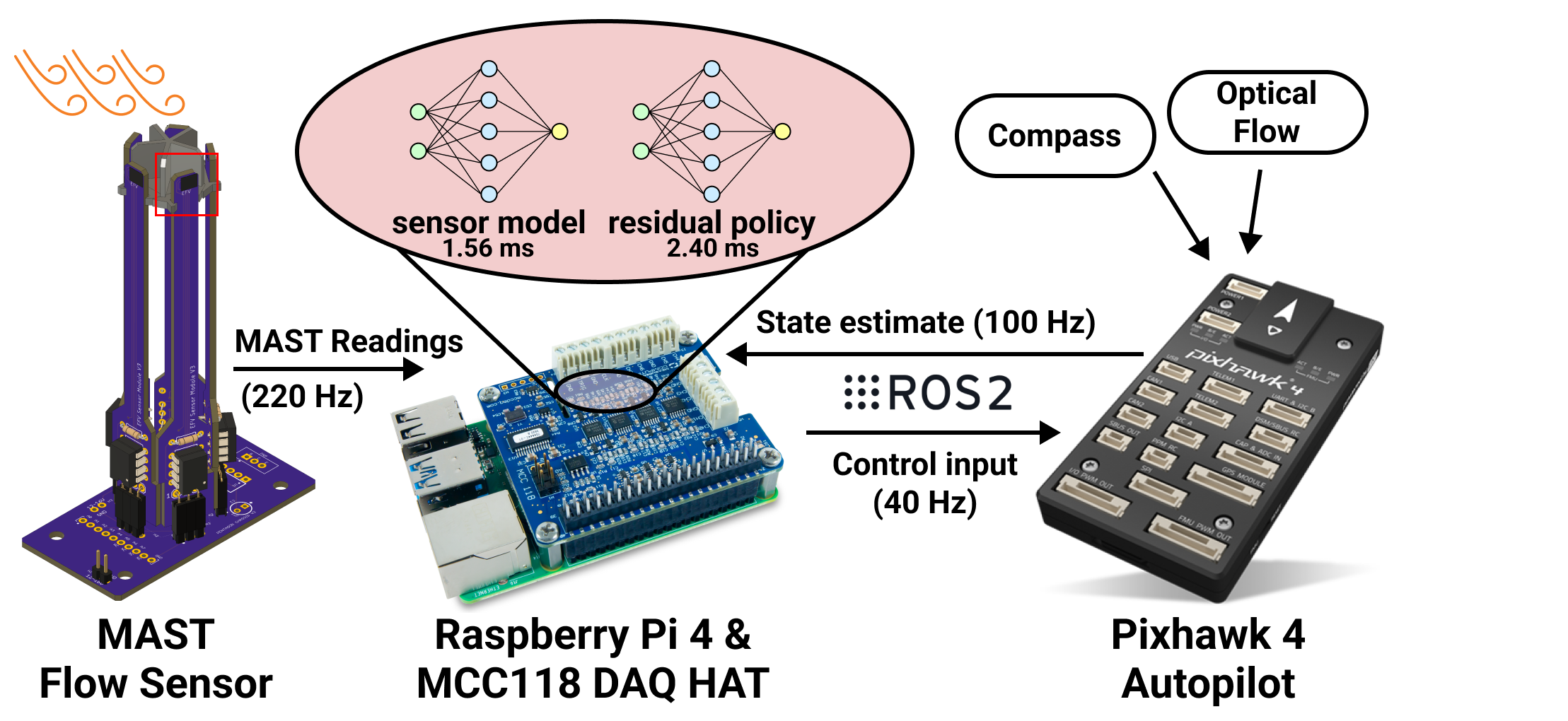}
\caption{System overview of the sensing and computing components. A red square highlights the MEMS Hotwire, shown in Fig. \ref{fig:efv}.}
\label{fig:system-overview}
\end{figure}

In this section, we provide an overview of the FlowDrone architecture and its main components; this architecture is illustrated in Fig.~\ref{fig:system-overview}. 

{\bf Drone hardware.} The FlowDrone platform is built on top of the Holybro X500 Kit  with a Pixhawk 4 autopilot. The drone is equipped with an accelerometer and magnetometer, as well as an onboard  Holybro M8N GPS for compass readings (the primary yaw reference), and a PX4FLOW Smart Camera and LIDAR-Lite v3 for optical flow and height measurements. The GPS is not used as the flights were conducted indoors. These sensors are used by the Pixhawk to calculate state estimates, which are then passed to an onboard companion computer (a Raspberry Pi 4 8~GB Model B) over the PX4-ROS2 bridge at 100~Hz. We highlight that all sensing and computation is \emph{onboard}, thus allowing for future outdoor deployment.

{\bf Sensors and wind estimation.} The key feature of the FlowDrone is the use of MEMS hot-wire technology for airflow sensing. These hot-wires are integrated in the MAST (MEMS anemometry sensing tower) to enable omnidirectional flow sensing. The operating principle of the sensors is described in Sec. \ref{sec:wind-sensing}. Raw sensor readings (in the form of voltages) are transmitted to the Raspberry Pi at 1~kHz using an MCC118 DAQ HAT. These raw measurements are then processed by a neural network running on the Raspberry Pi to compute an estimate of the wind magnitude and direction. This is described in Sec.~\ref{sec:sensor-model}. 

{\bf Wind-aware control.} The wind estimate and the drone's state estimate are utilized by a \emph{wind-aware controller} trained via reinforcement learning (RL) in simulation (with physical parameters estimated via system identification, and typical gust profiles measured experimentally). We describe the controller architecture and training process in Sec.~\ref{sec:wind-aware-control}. This controller operates at 40~Hz, publishing body rate and thrust setpoints to the Pixhawk 4. The Pixhawk 4 then executes rate control and mixing, before outputting actuator outputs to a power module, which distributes power and PWM signals to each motor. 

{\bf Experimental evaluation.} In order to evaluate the performance of the FlowDrone platform, the wind-aware controller is deployed with 5~m/s gusts generated by a fan array (Fig.~\ref{fig:anchor}). Our wind-aware controller's performance is compared to two strong baseline controllers, referred to as `baseline' and `wind-unaware.' The baseline controllers, hardware evaluation, and results are described in Sec.~\ref{sec:hardware-results}.

\begin{figure}
\centering
\includegraphics[width=0.30\textwidth]{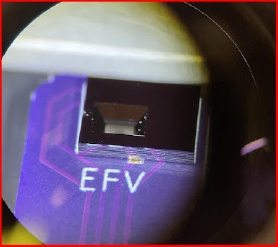}
\caption{An image of the MEMS hot-wire sensor, bonded to the vertical MAST PCBs, under microscope. This image corresponds to the red box in Fig. \ref{fig:system-overview}. Flow is measured as it passes through the window perpendicular to the PCB. The exposed microscale platinum ribbons lie recessed in the plane between the silicon and PCB; this provides some protection against accidental contact with the silicon. The MEMS hot-wires in this paper have been operational for 5+ months and even withstood vehicle crashes, a testament to the system's durability.}
\label{fig:efv}
\end{figure}

%% file: sections/sensor.tex
\section{Wind Sensing}\label{sec:wind-sensing}
In this section, we summarize our previous work \cite{mst} on designing the MAST: an omnidirectional flow sensor suitable for integration on multirotor UAVs. The MAST in this paper uses five hot-wire sensing elements arranged in a pentagonal configuration (Fig.~\ref{fig:anchor} inset).

\subsection{Sensing Element: the MEMS hot-wire}\label{sec:MEMS-hotwire}

Omnidirectional flow sensing on the MAST is enabled through the use of multiple MEMS hot-wire sensing elements. Each element consists of four platinum ribbons arranged in a Wheatstone bridge configuration on a silicon substrate using standard MEMS fabrication procedures. Wet-etch techniques are used to create a flow-through window on the silicon, making one of the four ribbons freestanding. The MEMS hot-wire bonded to a PCB is shown in Fig. \ref{fig:efv}. A 10~V constant voltage applied to the top of the bridge heats each ribbon; as air flows past the MEMS hot-wire and through the window, the rate of convective cooling differs between the exposed and embedded hot-wires. The resulting instantaneous change in temperature (and thus resistance) is measured as a change in the voltage across the bridge. Thereby, the change in voltage is directly related to a change in flow speed. In addition, we measured the bandwidth of the MEMS hot-wire through square-wave testing, applying a 10~Hz 0.1~V square wave to the top of the bridge to simulate a flow signal. An overdamped transfer function was fit to the 300 phase-averaged responses. The -3~dB bandwidth of the transfer function is 570~Hz, revealing that based on temporal response alone, the MEMS hot-wire is a stronger candidate for flow sensing in unsteady environments than pressure-based sensors, which can have settling times on the order of tens of seconds.

\subsection{MEMS Anemometry Sensing Tower (MAST)} \label{sec:MAST}

\begin{figure}
\centering
\includegraphics[width=0.48\textwidth]{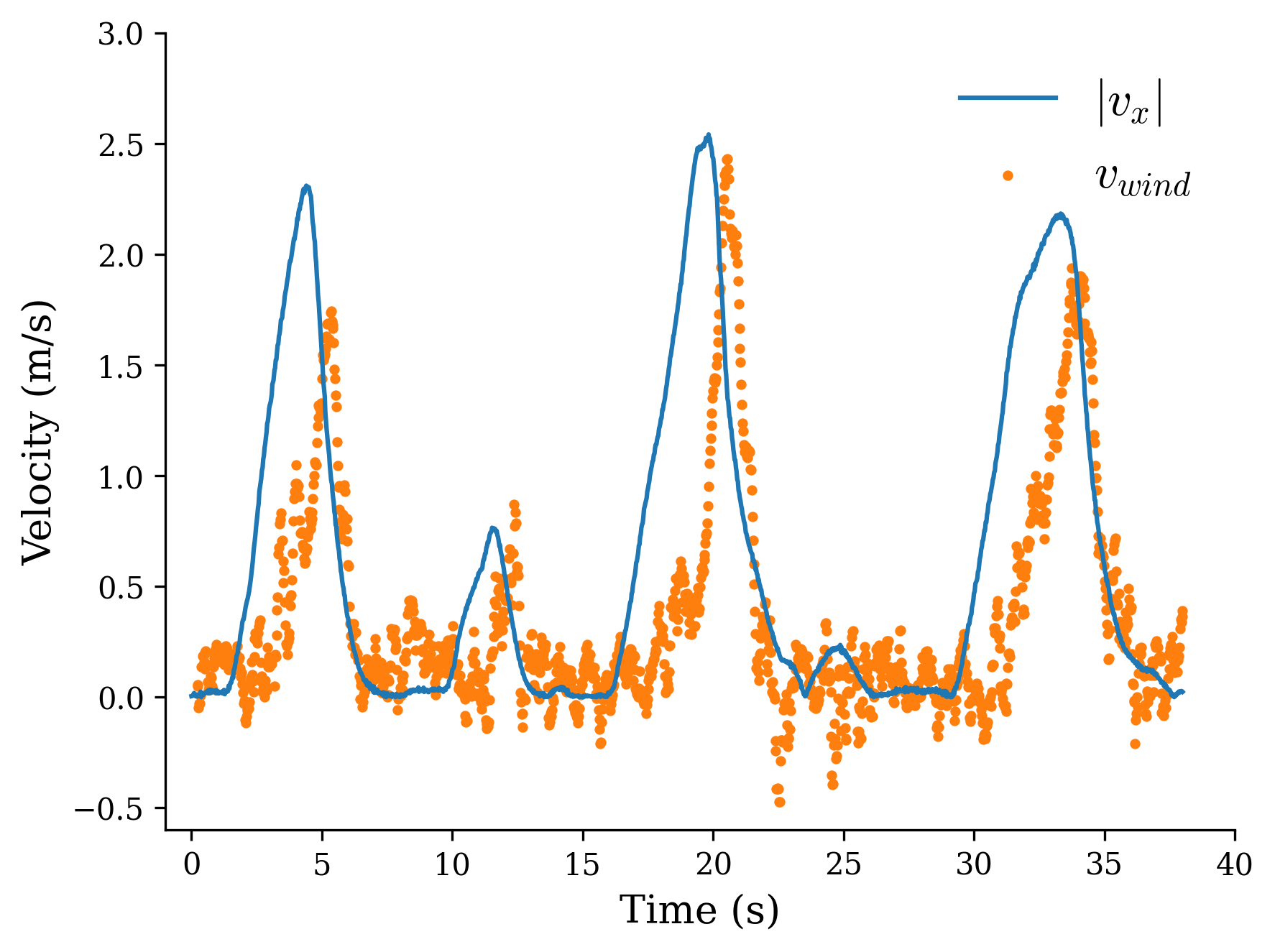}
\caption{UAV speed and estimated wind speed (relative wind) as the UAV flies approximately 8 meters back and forth in no-wind conditions in the flight arena. A rolling average over 10 samples is applied. The sensor measures the relative wind speed reasonably well with a slight delay that could be aerodynamic. This method is a proxy to validate the wind estimate, as there are other effects, e.g., the drone's angular velocity, which distinguish the drone's velocity from the flow at the MAST.}
\label{fig:relative-wind-speed}
\end{figure}

The MAST flow sensor consists of five MEMS hot-wire sensors attached to vertical printed circuit boards (PCBs), as shown in Fig.~\ref{fig:efv} and Fig.~\ref{fig:system-overview}. These PCBs project the MEMS hot-wires sufficiently above the rotor plane (150 mm) to measure the free-stream velocity. The MAST performance was validated through wind tunnel testing at velocities 1.3--5.0~m/s and orientations 0\textdegree and 360\textdegree, with accuracies reported in Sec. \ref{sec:sensor-model}. These tests suggested that a pentagonal geometry provides sufficient coverage to accurately measure a two-dimensional wind vector. Additionally, Fig. \ref{fig:relative-wind-speed} compares the MAST wind estimate (of relative wind) to the vehicle speed (in still air), further validating its performance onboard the UAV.

\subsection{Sensor Model}
\label{sec:sensor-model}
A sensor model running on board the Raspberry Pi estimates the wind vector (direction and magnitude) from raw sensor readings. Due to the nonlinearity of the sensor, we trained a neural network to approximate the sensor-to-wind mapping from wind tunnel data. Separate networks were developed for wind direction and wind magnitude predictions.

The angle prediction network is a fully connected network with two hidden layers using a ReLU activation, while the speed prediction network is fully connected with one hidden layer, utilizing a Tanh activation. The MAST achieves the following performance in testing: $1.6^\circ$ expected angle prediction error (with an empirical 95\% error upper bound of $5.0^\circ$), and 0.14~m/s expected speed prediction error (with an empirical 95\% error upper bound of 0.36~m/s). This performance compares favorably to existing UAV flow-sensing methods. For example, the differential pressure probe techniques of \cite{yeo2015onboard, bruschi2016wind} achieve 15\% full-scale deflection and 12$^\circ$ mean errors, respectively. The approach in \cite{tagliabue2020touch} achieves root mean square error of $0.38$ m/s in velocity prediction, but lacks description of temporal resolution. Finally, \cite{prudden2017anemometer} is likely the closest comparison, achieving 1~m/s and 5$^\circ$ error at 400Hz bandwidth, at the expense (as noted before) of requiring a large, bulky sensor. Thus, the low-weight MAST and associated sensor model provide low-latency wind estimates to the control architecture that are as accurate as any existing method implementable on UAVs. 

%% file: sections/control.tex
\vspace{3pt}
\section{Wind-Aware Control}\label{sec:wind-aware-control}

\subsection{Simulated environment with wind} Since it is challenging and time-consuming to perform hardware-in-the-loop training of a controller in diverse wind conditions in a real lab setup, we train the policy in a simulated environment built upon \texttt{gym-pybullet-drones} \cite{panerati2021gymdrone}, an open-source drone simulation environment based on the PyBullet simulator \cite{Coumans18}. The lift force generated by the rotors is calculated based on thrust calibration results on the real drone. We also use physical parameters including mass and moment of inertia obtained via system identification and rely on the physics engine for dynamics simulation.

The effect of wind is simulated using the model from \cite{craig2020geometric}, which relates the wind speed and direction to the bluff-body and induced drag acting on the drone. We then manually add the drag forces in simulation at each timestep given the drone state. The simulated wind is generated in the positive X direction in the world frame. We vary the wind speed in a step-like profile (mimicking the measurements taken in real wind conditions as in Fig.~\ref{fig:x-position-comparison-with-wind} bottom), which consists of three stages: ``Low", ``Slope", and ``High". The wind speed and duration of the three stages are all randomized for training a robust wind-aware policy. An additional dip of the wind speed is added to the ``High" stage to mimic sudden instability, which was frequently observed during real flights as the flow generated by the fan array is unsteady and not spatially uniform. 

\subsection{Residual policy for wind compensation} \label{sec:residual-policy}
\begin{figure}
\centering
\includegraphics[width=0.48\textwidth]{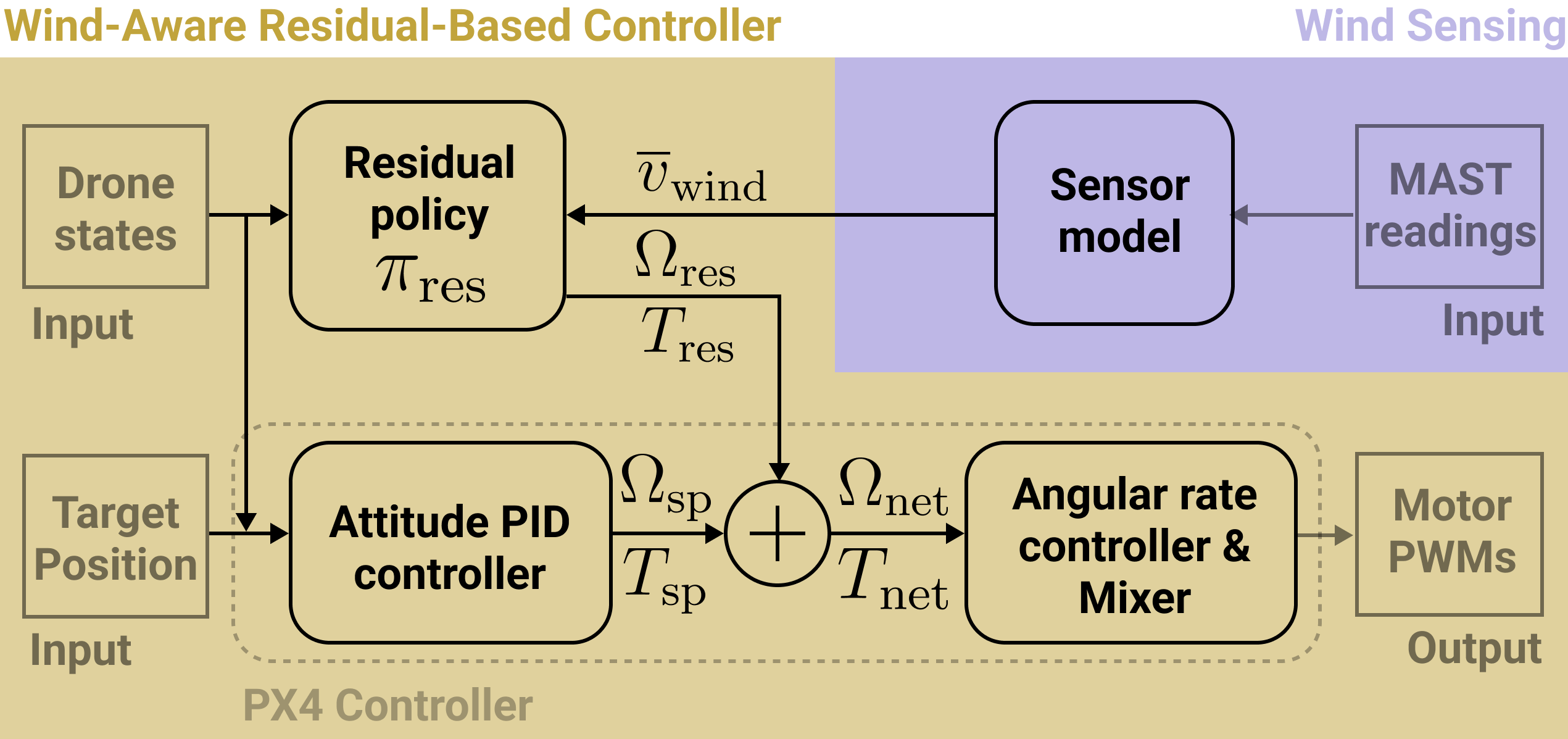}
\caption{Controller diagram showing the wind sensing and the wind-aware residual-based controller. Taking MAST voltage readings and drone states as input, the controller computes motor PWM commands compensating for wind disturbances.}
\label{fig:control-diagram}
\end{figure}

In order to perform wind compensation, we train a residual control policy on top of the open-source PX4 attitude controller (see Fig.~\ref{fig:control-diagram} for a visualization of the overall controller architecture). Using a reinforcement learning approach (instead of a model-based approach) allows us to train a nonlinear policy which can potentially leverage temporal structure in wind gusts. The trained residual policy takes real-time wind estimates and drone states as input and outputs additional body angular rates $\Omega_\text{res}$ and thrust $T_\text{res}$, which are then added to the respective setpoints $\Omega_\text{sp}, T_\text{sp}$ calculated by the upstream PX4 attitude PID controller. The net setpoints $\Omega_\text{net}, T_\text{net}$ are then fed into the downstream angular rate controller and mixer:
\vspace{1pt}
\begin{align}
\Omega_\text{net} & = \Omega_\text{sp} + \Omega_\text{res}, \\
T_\text{net} & = T_\text{sp} + T_\text{res}.
\end{align}
\vspace{1pt}
The overall controller runs at 40 Hz. The residual policy $\pi_\text{res}$ is parameterized as a multi-layer perceptron (MLP) with hidden layer sizes $[512, 256, 128, 128]$ and ReLU activation. The input to the residual policy
\vspace{1pt}
\begin{equation}
    [\Omega_\text{res}, T_\text{res}] = \pi_\text{res}(r, \Psi, v, w, \overline{v}_\text{wind})
\end{equation}
\vspace{1pt}
includes the drone's current 3D position ($r = [x, y, z]$), orientation in roll, pitch, and yaw ($\Psi = [\phi, \theta, \psi]$), linear velocity ($v = \dot{r}$), and angular velocity ($w = [w_x, w_y, w_z]$), all relative to the world frame. In addition, the residual policy takes as input the wind measurement at the current timestep $t$ and past four timesteps (only the components in the X direction of the world frame). We skip $t_\text{s} = 5$ steps between each wind measurement ($\overline{v}_\text{wind} = [v_\text{wind}^t, v_\text{wind}^{t-t_\text{s}}, v_\text{wind}^{t-2t_\text{s}}, v_\text{wind}^{t-3t_\text{s}}, v_\text{wind}^{t-4t_\text{s}}]$). Since the control loop is 40~Hz, this roughly covers a time window of 0.5~s. The outputs, $\Omega_\text{res}$ and $T_\text{res}$, are normalized between $[-0.3, 0.3]$~rad/s and $[-1,1]$~N respectively using the Tanh activation function. 

On the real hardware platform, the wind measurements are obtained by processing the MAST readings through the sensor model described in Sec.~\ref{sec:sensor-model}. In simulation, we treat the wind sensor model as perfect, meaning the wind measurement is the same as the simulated wind. Since the wind profile contains a non-trivial amount of noise, we filter the measurements using a rolling maximum over the past 0.1~s both in simulation and hardware.

We find that large values of body rate setpoints $\Omega_\text{sp}$ from the attitude controller can hinder the training progress of the residual policy. Thus we clip $\Omega_\text{sp}$ to be in $[-0.1, 0.1]$~rad/s for the wind-aware controller. 

\subsection{Reinforcement learning training} We train the residual policy in simulation using Soft Actor Critic \cite{haarnoja2018sac}, an off-policy reinforcement learning algorithm. The task of the drone is to hover at the target position $[0,0,1]$~m in an inertial East-North-Up frame for a 10-second horizon (thus 400 timesteps using 40 Hz control rate). The reward function is defined as the negative of the distance of the current drone position to the target. We randomize the initial position and orientation of the drone at each rollout; the ranges of the initial 3D positions, roll and pitch angles, and yaw angle are $[-30,30]$~cm, $[-0.1, 0.1]$~rad, and $[-0.3, 0.3]$~rad. The model is trained with 10 million total simulation timesteps.

%% file: sections/results.tex
\section{Hardware Experiments}\label{sec:hardware-results}

In order to demonstrate the effectiveness of MEMS hot-wire sensors for UAV control, we evaluated the wind-aware controller's performance in tracking a hover setpoint in the presence of a wind gust (same task as in simulation) in a real setting. We compare the following three controllers:
\begin{itemize}[leftmargin=*]
    \item \textbf{Wind-Aware Residual-Based Controller (``wind-aware'').} This is our controller as described in Sec.~\ref{sec:wind-aware-control} and Fig.~\ref{fig:control-diagram}.
    \item \textbf{Wind-Unaware Residual-Based Controller (``wind-unaware'').} This controller has the same architecture and is trained with the same conditions as wind-aware, except that the residual policy does not have access to the wind estimate $\bar{v}_{wind}$. Differences in performance between this controller and the wind-aware controller thus directly provide evidence for the benefits of utilizing wind measurements for control. 

    \item \textbf{PX4 Attitude Controller (``baseline'').} We use the popular open-source PX4 Autopilot for attitude control, publishing body rate and thrust setpoints to the PX4 at 40~Hz over ROS2. To integrate with the learning framework, the attitude controller has been adapted into Python \cite{bobzwik}. Referencing Fig.~\ref{fig:control-diagram}, this controller is the ``PX4 controller'' (the bottom half without wind sensing or a residual policy). Unlike wind-aware and wind-unaware, the body rate setpoint $\Omega_\text{sp}$ is not clipped here.
\end{itemize}

\subsection{Experiment setup} 

We conducted 10 flights for each of the three controllers in controlled gust conditions. The \href{https://www.youtube.com/watch?v=KWqkH9Z-338}{supplementary video} \cite{video} demonstrates representative trials of all three controllers. We used six high velocity (350 cfm) blowers to generate the gusts during hardware evaluation. The blowers were arranged in a $2\times3$ array (Fig.~\ref{fig:anchor}), with the top row blowers inverted, generating a flow volume of $22\times86$~cm at the blower exit. By the time the flow volume reaches the drone, positioned 4~m away from the blower outlets, the flow volume encompasses the entire drone for the duration of the flight test. The peak gust speed was approximately 5~m/s (verified by a handheld anemometer) for all experiments. It should be noted that gaps between the fans and inherent turbulence resulted in a spatio-temporally-varying flow.

For each test, the drone was commanded to take off and hover at $r_{sp}=[0,0,1]$~m, $\Psi_{sp}=[0,0,0]$~rad, in an inertial East-North-Up frame. This remained the setpoint for the duration of the flight. Once in position, the drone switches to the desired controller: baseline, wind-unaware, or wind-aware. From $t=[0,12)$~s, the drone continues to hover with zero wind. This delay (which was not simulated during training) is used to ensure that the residual policy has not learned an open-loop prediction of when the gust will start and that it is capable of flying in no-wind conditions. At $t=12$~s, the fans are turned on to their maximum setting for the remaining $18$~s of flight ($T = 30$~s total). The fans are oriented to blow in the $+X$ (i.e. East) direction of the inertial frame, which at $\psi = 0$ is at the back of the drone, matching simulation conditions.

\subsection{Results and Discussions} 

\begin{figure*}[t]
\centering
\includegraphics[width=0.96\textwidth]{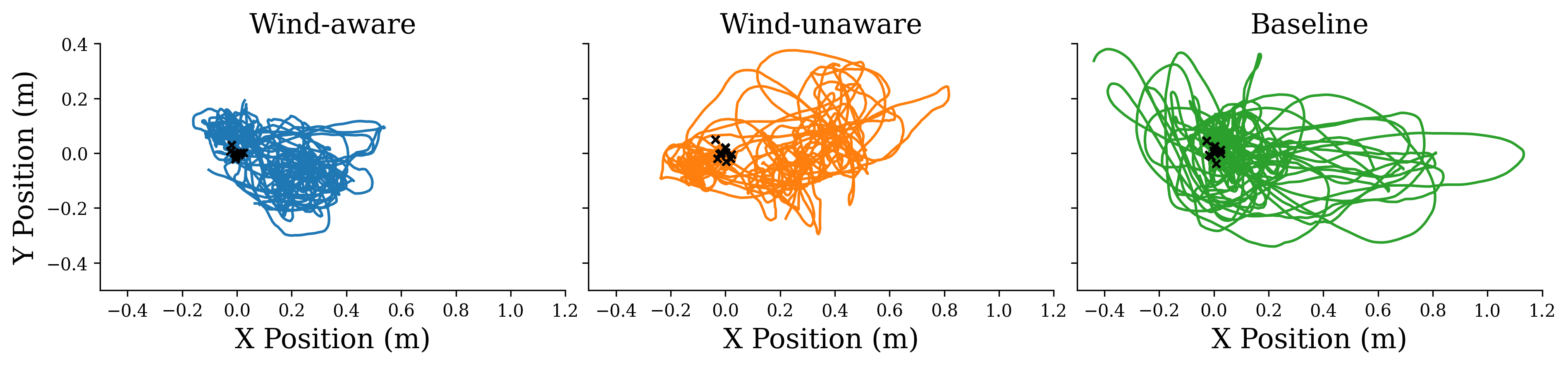}
\caption{X and Y trajectories of all ten trials for each of the three controllers (wind-aware, wind-unaware, and baseline) while attempting to maintain a hover position throughout a 0-5~m/s gust in the $+X$ direction. The wind-aware controller outperforms the two baselines. Here, we measure performance as the ability to reduce both maximum and overall error in the presence of wind. Starting positions for each trial are marked by black $\times$s.}
\label{fig:xy-plots}
\end{figure*}

\begin{figure}[ht]
\centering
\includegraphics[width=0.48\textwidth]{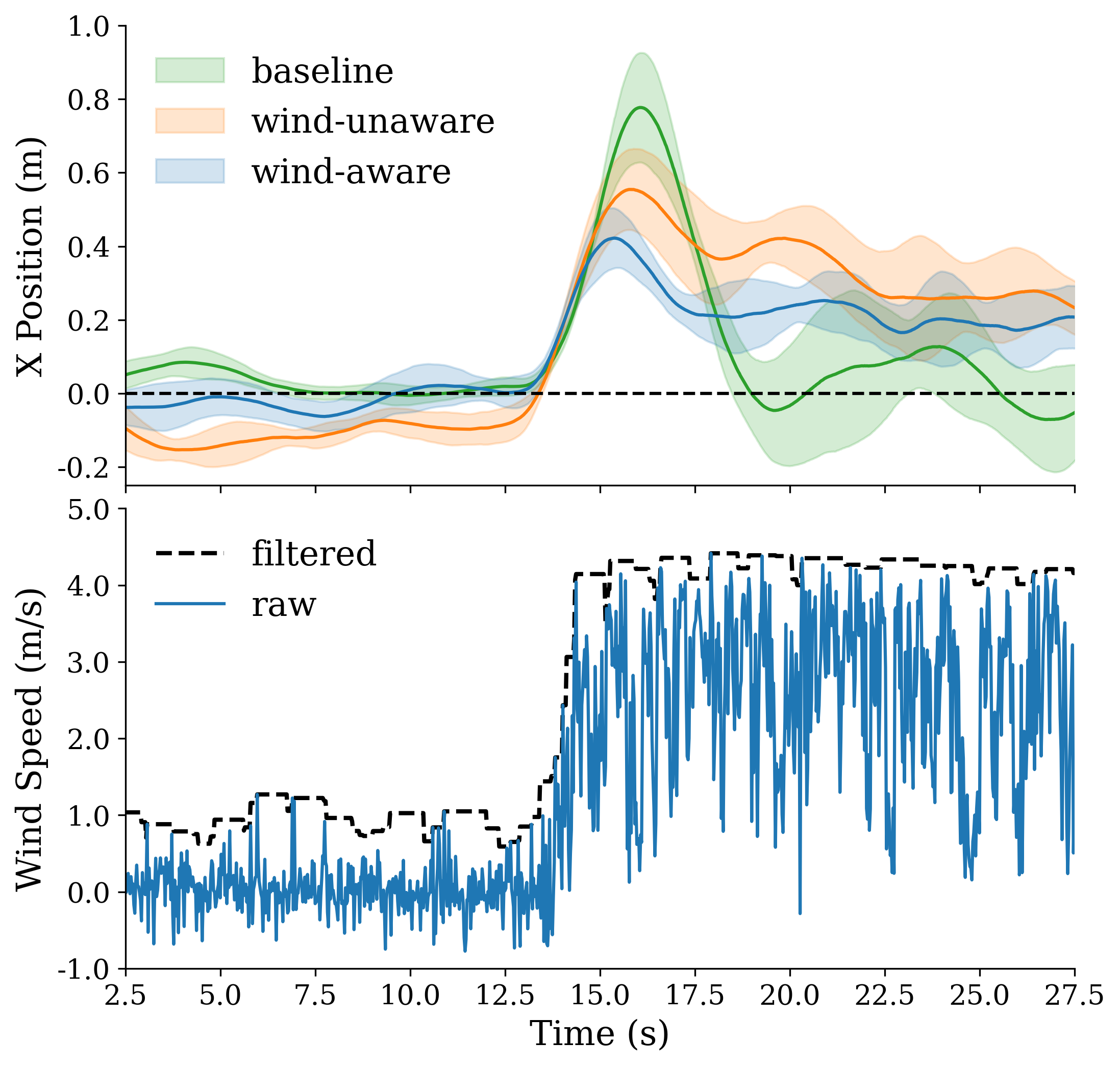}
\caption{(Top) Average $X$ trajectory (line) for each controller over ten runs with associated pointwise standard deviation (shaded). Trajectories were shifted (within $ \pm 0.75$s) to align the gust onset. (Bottom) Raw and filtered wind estimates for a representative gust measured during a wind-aware flight. The filter effectively reduces high-frequency noise in the flow and captures the gust onset.
}
\label{fig:x-position-comparison-with-wind}
\end{figure}

The 10 trajectories for each controller are plotted in Fig.~\ref{fig:xy-plots}. Qualitatively, the wind-aware trajectory is more concentrated, with visibly less maximum error and variance in both $X$ and $Y$. In Fig.~\ref{fig:x-position-comparison-with-wind} (top), the average $X$-trajectory of each controller is plotted against time, with bands for $\pm1$ standard deviation. Again, the wind-aware controller deviates the least from the setpoint. We observe that the average baseline trajectory returns to $x=0$ faster than both residual-based controllers. This is a topic for future investigation; however, one hypothesis is that extending the simulation episode length beyond 10 seconds (where the gust duration was 4-6 seconds), will improve performance in hardware (where the episode length is 30 seconds, and the gust duration is 15 seconds).

In Fig.~\ref{fig:x-position-comparison-with-wind} (bottom), we plot a representative gust measured during a wind-aware flight. 
The raw wind estimate is shown, as well as the output of a moving maximum filter applied over the latest 100 wind estimates (0.1 seconds), which represents the estimate provided to the residual policy in Sec. \ref{sec:residual-policy}.

The performance of each controller is evaluated by three metrics on the $X$ trajectory: maximum error (or deviation from the hover setpoint), mean-squared error (over the entire trajectory), and total range (the difference between the maximum and minimum) across the trajectory. The results are shown in Table \ref{table:RunStats}. Max error most directly penalizes the gust onset effect, while mean squared error (MSE) is taken over the entire trajectory. The range metric additionally penalizes under- or over-shoot. By each metric, the wind-aware controller outperforms the others; this illustrates the wind-aware controller's ability to reduce both maximal and overall error in the presence of wind. Specifically, in terms of the maximum error metric, the wind-aware controller performance improves (on average over the 10 runs) by 44\% over the baseline controller and by 24\% over the wind-unaware controller.

\begin{table}[h!]
\begin{center}
\begin{tabular}{ c || c c c } 
& Wind-aware & Wind-unaware & Baseline \\
\hhline{=|===}
 Max Error (m) & $\mathbf{0.441}$ $(0.064)$&  $0.582$ $(0.094)$& $0.780$ $(0.142)$\\ \hhline{-|   }
 MSE (m$^2$) &  $\mathbf{0.035}$ $(0.006)$& $0.079$ $(0.013)$& $0.057$ $(0.016)$ \\ \hhline{-|   }
 Range (m) & $\mathbf{0.538}$ $(0.072)$& $0.773$ $(0.100)$& $0.962$ $(0.222)$\\
\end{tabular}
\end{center}
\caption{Hardware performance of each controller along several metrics, with the standard deviation in parentheses. The minimum entry of each row represents the best performance and is denoted by boldface font.}
\label{table:RunStats}
\end{table}

%% file: sections/conclusion.tex
\section{Conclusions and Future Work}

We have presented the FlowDrone: a multirotor UAV platform that integrates fast-response hot-wire sensors for real-time wind estimation. These new MEMS-based sensors afford several benefits over traditional methods for sensing wind including low latency, a small form factor, and high accuracy. We implemented a reinforcement learning pipeline to leverage the MEMS hot-wire sensors for gust rejection on the FlowDrone. We compared the FlowDrone's wind-aware controller with two strong wind-unaware baselines in a setting with 5 m/s wind gusts. Over 30 total flights, we demonstrated significant improvements in tracking a hover setpoint and the importance of direct wind measurements. 

In future work, we plan to vary gust direction as well as magnitude. With the FlowDrone platform and RL pipeline ready, this benefit should be realizeable just by varying the wind direction in simulation during the training of the wind-aware and wind-unaware controllers. Additionally, we plan to investigate optimal sensor placement for improving control performance. Currently, the sensors are placed on the MAST in order to estimate the free-stream wind velocity. However, sensors placed in proximity to the propellers may also provide information that helps estimate effects of wind on the propeller's aerodynamics. Another direction of practical relevance is further miniaturization of the PCB circuitry through surface-mount technology to deploy MAST on `nano' multirotor platforms (e.g., Crazyflie 2.1). Finally, a particularly exciting direction is to utilize the MEMS hot-wire sensors for improving drone control close to surfaces (e.g., ground and obstacles) by measuring local airflow patterns due to ground and surface effects.

\newpage